\documentclass[runningheads,a4paper]{llncs}

\usepackage{amsfonts, amssymb, amsmath}
\usepackage{float}
\usepackage{scrextend}

\usepackage{graphicx}

\usepackage{multirow}
\usepackage{booktabs}

\usepackage{hyperref}

\newcommand{\keywords}[1]{\par\addvspace\baselineskip
\noindent\keywordname\enspace\ignorespaces#1}

\renewcommand\refname{}

\newcommand{\norm}[1]{\left\lVert#1\right\rVert}

\usepackage{xcolor}

\definecolor{nice-green}{HTML}{4DAF4A}
\definecolor{nice-blue}{HTML}{377EB8}

\begin{document}

\mainmatter  

\title{Implementing Neural Turing Machines}

\titlerunning{Implementing Neural Turing Machines}

\author{
	Mark Collier \inst{1} 
	\and 
	Joeran Beel \inst{1}
	}

\institute{
Trinity College Dublin\\
\email{[mcollier, joeran.beel]@tcd.ie}}

\maketitle              

\begin{abstract}

Neural Turing Machines (NTMs) are an instance of Memory Augmented Neural Networks, a new class of recurrent neural networks which decouple computation from memory by introducing an external memory unit. NTMs have demonstrated superior performance over Long Short-Term Memory Cells in several sequence learning tasks. A number of open source implementations of NTMs exist but are unstable during training and/or fail to replicate the reported performance of NTMs. This paper presents the details of our successful implementation of a NTM. Our implementation learns to solve three sequential learning tasks from the original NTM paper. We find that the choice of memory contents initialization scheme is crucial in successfully implementing a NTM. Networks with memory contents initialized to small constant values converge on average 2 times faster than the next best memory contents initialization scheme.


\end{abstract}
\vspace{-1cm}

\section{Introduction} \label{introduction}

Neural Turing Machines (\textbf{NTMs}) \cite{RN11} are one instance of several new neural network architectures \cite{RN11,RN12,RN38} classified as Memory Augmented Neural Networks (\textbf{MANNs}). MANNs defining attribute is the existence of an external memory unit. This contrasts with gated recurrent neural networks such as Long Short-Term Memory Cells (\textbf{LSTMs}) \cite{RN18} whose memory is an internal vector maintained over time. LSTMs have achieved state-of-the-art performance in many commercially important sequence learning tasks, such as handwriting recognition \cite{RN36}, machine translation \cite{RN16} and speech recognition \cite{RN37}. But, MANNs have been shown to outperform LSTMs on several artificial sequence learning tasks that require a large memory and/or complicated memory access patterns, for example memorization of long sequences and graph traversal \cite{RN11,RN12,RN52,RN38}.

The authors of the original NTM paper, did not provide source code for their implementation. Open source implementations of NTMs exist\footnote{\label{first_impl}https://github.com/snowkylin/ntm}\footnote{\label{second_impl}https://github.com/chiggum/Neural-Turing-Machines}\footnote{\label{third_impl}https://github.com/yeoedward/Neural-Turing-Machine}\footnote{\label{fourth_impl}https://github.com/loudinthecloud/pytorch-ntm}\footnote{\label{fifth_impl}https://github.com/camigord/Neural-Turing-Machine}\footnote{\label{sixth_impl}https://github.com/snipsco/ntm-lasagne}\footnote{\label{seventh_impl}https://github.com/carpedm20/NTM-tensorflow} but a number of these implementations\footref{fifth_impl}\footref{sixth_impl}\footref{seventh_impl} report that the gradients of their implementation sometimes become NaN during training, causing training to fail. While others report slow convergence or do not report the speed of learning of their implementation. The lack of a stable open source implementation of NTMs makes it more difficult for practitioners to apply NTMs to new problems and for researchers to improve upon the NTM architecture.

In this paper we define a successful NTM implementation\footnote{\label{source_code}Source code at: \url{https://github.com/MarkPKCollier/NeuralTuringMachine}} which learns to solve three benchmark sequential learning tasks \cite{RN11}. We specify the set of choices governing our NTM implementation. We conduct an empirical comparison of a number of memory contents initialization schemes identified in other open source NTM implementations. We find that the choice of how to initialize the contents of memory in a NTM is a key factor in a successful NTM implementation. We base our Tensorflow implementation on another open source NTM implementation\footref{first_impl}, but following our experimental results make significant changes to the memory contents initialization, controller head parameter computation and interface which result in faster convergence, more reliable optimization and easier integration with existing Tensorflow methods. Our implementation is available publicly under an open source license\footref{source_code}.

\section{Neural Turing Machines} \label{neural_turing_machines}

NTMs consist of a controller network which can be a feed-forward neural network or a recurrent neural network and an external memory unit which is a $N * W$ memory matrix, where $N$ represents the number of memory locations and $W$ the dimension of each memory cell. Whether the controller is a recurrent neural network or not, the entire architecture is recurrent as the contents of the memory matrix are maintained over time. The controller has read and write heads which access the memory matrix. The effect of a read or write operation on a particular memory cell is weighted by a soft attentional mechanism. This addressing mechanism is similar to attention mechanisms used in neural machine translation \cite{RN6,RN10} except that it combines location based addressing with the content based addressing found in these attention mechanisms.

In particular for a NTM, at each timestep ($t$), for each read and write head the controller outputs a set of parameters: $\mathbf{k}_t$, $\beta_t \geq 0$, $g_t \in [0, 1]$, $\mathbf{s}_t$ (s.t. $\sum_k s_t(k) = 1$ and $\forall k \ s_t(k) \geq 0$) and $\gamma_t \geq 1$ which are used to compute the weighting $\mathbf{w}_t$ over the N memory locations in the memory matrix $\mathbf{M}_t$ as follows:

\begin{equation} \label{eq:content_based_addr}
w^c_t(i) \gets \frac{exp(\beta_t K[\mathbf{k}_t, \mathbf{M}_t(i)])}{\sum_{j=0}^{N-1} exp(\beta_t K[\mathbf{k}_t, \mathbf{M}_t(j)])}
\end{equation}

$\mathbf{w}^c_t$ allows for content based addressing where $\mathbf{k}_t$ represents a lookup key into memory and $K$ is some similarity measure such as cosine similarity:

\begin{equation} \label{eq:cosine_similarity}
K[\mathbf{u}, \mathbf{v}] = \frac{\mathbf{u} \cdot \mathbf{v}}{\norm{\mathbf{u}} \cdot \norm{\mathbf{v}}}
\end{equation}

Through a series of operations NTMs also enable iteration from current or previously computed memory weights as follows:

\begin{equation} \label{eq:interpolation}
\mathbf{w}^{g}_{t} \gets g_t \mathbf{w}^c_t + (1 - g_t) \mathbf{w}_{t-1}
\end{equation}

\begin{equation} \label{eq:conv_shift_eq}
\tilde{w}_{t}(i) \gets \sum_{j=0}^{N-1} w^{g}_{t}(j) s_t(i-j)
\end{equation}

\begin{equation} \label{eq:sharpening}
w_{t}(i) \gets \frac{\tilde{w}_{t}(i)^{\gamma_t}}{\sum_{j=0}^{N-1} \tilde{w}_{t}(j)^{\gamma_t}}
\end{equation}

\noindent where (\ref{eq:interpolation}) enables the network to choose whether to use the current content based weights or the previous weight vector, (\ref{eq:conv_shift_eq}) enables iteration through memory by convolving the current weighting by a 1-D convolutional shift kernel and (\ref{eq:sharpening}) corrects for any blurring occurring as a result of the convolution operation.

The vector $\mathbf{r}_t$ read by a particular read head at timestep $t$ is computed as:

\begin{equation} \label{eq:read}
\mathbf{r}_t \gets \sum_{i=0}^{N-1} w_t(i) \mathbf{M}_t(i)
\end{equation}

Each write head modifies the memory matrix at timestep $t$ by outputting additional erase ($\mathbf{e}_t$) and add ($\mathbf{a}_t$) vectors:

\begin{equation} \label{eq:write_erase}
\tilde{\mathbf{M}}_t(i) \gets \mathbf{M}_{t-1}(i)[\mathbf{1} - w_t(i) \mathbf{e}_t]
\end{equation}

\begin{equation} \label{eq:write_add}
\mathbf{M}_t(i) \gets \tilde{\mathbf{M}}_t(i) + w_t(i) \mathbf{a}_t
\end{equation}

Equations (\ref{eq:content_based_addr}) to (\ref{eq:write_add}) define how addresses are computed and used to read and write from memory in a NTM, but many implementation details of a NTM are open to choice. In particular the choice of the similarity measure $K$, the initial weightings $\mathbf{w}_0$ for all read and write heads, the initial state of the memory matrix $\mathbf{M}_0$, the choice of non-linearity to apply to the parameters outputted by each read and write head and the initial read vector $\mathbf{r}_0$ are all undefined in a NTM's specification.

While any choices for these satisfying the constraints on the parameters outputted by the controller would be a valid NTM, in practice these choices have a significant effect on the ability of a NTM to learn.

\section{Our Implementation} \label{implementation}

\noindent \textbf{Memory contents initialization} - We hypothesize that how the memory contents of a NTM are initialized may be a defining factor in the success of a NTM implementation. We compare the three memory contents initialization schemes that we identified in open source implementations of NTMs. In particular, we compare \textit{constant initialization} where all memory locations are initialized to $10^{-6}$, \textit{learned initialization} where we backpropagate through initialization and \textit{random initialization} where each memory location is initialized to a value drawn from a truncated Normal distribution with mean $0$ and standard deviation $0.5$. We note that five of the seven implementations\footref{first_impl}\footref{second_impl}\footref{third_impl}\footref{fourth_impl}\footref{fifth_impl} we identified randomly initialize the NTM's memory contents. We also identified an implementation which initialized memory contents to a small constant value\footref{sixth_impl} and an implementation where the memory initialization was learned\footref{seventh_impl}.

\textit{Constant initialization} has the advantage of requiring no additional parameters and providing a stable known memory initialization during inference. \textit{Learned initialization} has the potential advantage of learning an initialization that would enable complex non-linear addressing schemes \cite{RN52} while also providing stable initialization after training. This comes at the cost of $N * W$ extra parameters. \textit{Random initialization} has the potential advantage of acting as a regularizer, but it is possible that during inference memory contents may be in a space not encountered during training.

\textbf{Other parameter initialization} - Instead of initializing the previously read vectors $\mathbf{r}_0$ and address weights $\mathbf{w}_0$ to bias values as per \cite{RN11} we backpropagate through their initialization and thus initialize them to a learned bias vector. We argue that this initialization scheme provides sufficient generality for tasks that require more flexible initialization with little cost in extra parameters (the number of additional parameters is $W * H_r + N * (H_r + H_w)$ where $H_r$ is the number of read heads and $H_w$ is the number of write heads). For example, if a NTM with multiple write heads wishes to write to different memory locations at timestep $1$ using location based addressing then $\mathbf{w}_0$ must be initialized differently for each write head. Having to hard code this for each task is an added burden on the engineer, particularly when the need for such addressing may not be known a priori for a given task, thus we allow the network to learn this initialization.

\textbf{Similarity measure} - For $K$, we follow \cite{RN11} in using  cosine similarity (\ref{eq:cosine_similarity}) which scales the dot product into the fixed range $[-1, 1]$.

\textbf{Controller inputs} - At each timestep the controller is fed the concatenation of the input coming externally into the NTM $\mathbf{x}_t$ and the previously read vectors $\mathbf{r}_{t-1}$ from all of the read heads of the NTM. We note that such a setup has achieved performance gains for attentional encoder-decoders in neural machine translation \cite{RN10}.

\textbf{Parameter non-linearities} - Similarly to a LSTM we force the contents of the memory matrix to be in the range $[-1, 1]$, by applying the $\tanh$ function to the outputs of the controller corresponding to $\mathbf{k}_t$ and $\mathbf{a}_t$ while we apply the sigmoid function to the corresponding erase vector $\mathbf{e}_t$. We apply the function $softplus(x) \gets \log(\exp(x) + 1)$ to satisfy the constraint $\beta_t \geq 0$. We apply the logistic sigmoid function to satisfy the constraint $g_t \in [0, 1]$. In order to make the convolutional shift vector $\mathbf{s}_t$ a valid probability distribution we apply the softmax function. In order to satisfy $\gamma_t \geq 1$ we first apply the $softplus$ function and then add $1$.

\section{Methodology} \label{methodology}

\subsection{Tasks}

We test our NTM implementation on three of the five artificial sequence learning tasks described in the original NTM paper \cite{RN11}.

\textbf{Copy} - for the Copy task, the network is fed a sequence of random bit vectors followed by an end of sequence marker. The network must then output the input sequence. This requires the network to store the input sequence and then read it back from memory. In our experiments we train and test our networks on 8-bit random vectors with sequences of length sampled uniformly from $[1, 20]$.

\textbf{Repeat Copy} - similarly to the Copy task, with Repeat Copy the network is fed an input sequence of random bit vectors. Unlike the Copy task, this is followed by a scalar that indicates how many times the network should repeat the input sequence in its output sequence. We train and test our networks on 8-bit random vectors with sequences of length sampled uniformly from $[1, 10]$ and number of repeats also sampled uniformly from $[1, 10]$.

\textbf{Associative Recall} - Associative Recall is also a sequence learning problem with sequences consisting of random bit vectors. In this case the inputs are divided into items, with each item consisting of 3 x 6-dimensional vectors. After being fed a sequence of items and an end of sequence marker, the network is then fed a query item which is an item from the input sequence. The correct output is the next item in the input sequence after the query item. We train and test our networks on sequences with the number of items sampled uniformly from $[2, 6]$.

\subsection{Experiments}

We first run a set of experiments to establish the best memory contents initialization scheme. We compare the \textit{constant}, \textit{random} and \textit{learned} initialization schemes on the above three tasks. We demonstrate below (Sec. \ref{results}) that the best such scheme is the \textit{constant initialization} scheme. We then compare the NTM implementation described above (Sec. \ref{implementation}) under the \textit{constant initialization} scheme to two other architectures on the Copy, Repeat Copy and Associative Recall tasks. We follow the NTM authors \cite{RN11} in comparing our NTM implementation to a LSTM network. As no official NTM implementation has been made open source, as a further benchmark, we compare our NTM implementation to the official implementation\footnote{\label{dnc}https://github.com/deepmind/dnc} of a Differentiable Neural Computer (\textbf{DNC}) \cite{RN12}, a successor to the NTM. This provides a guide as to how a stable MANN implementation performs on the above tasks.

In all of our experiments for each network we run training 10 times from different random initializations. To measure the learning speed, every 200 steps during training we evaluate the network on a validation set of 640 examples with the same distribution as the training set.

For all tasks the MANNs had 1 read and 1 write head, with an external memory unit of size 128 X 20 and a LSTM controller with 100 units. The controller outputs were clipped elementwise to the range $(-20, 20)$. The LSTM networks were all a stack of 3 X 256 units. All networks were trained with the Adam optimizer \cite{RN15} with learning rate 0.001 and on the backward pass gradients were clipped to a maximum gradient norm of 50 as described in \cite{pascanu2013difficulty}.

\section{Results} \label{results}

\subsection{Memory Initialization Comparison}

We hypothesized that how the memory contents of a NTM were initialized would be a key factor in a successful NTM implementation. We compare the three memory initialization schemes we identified in open source NTM implementations. We then use the best identified memory contents initialization scheme as the default for our NTM implementation.

\begin{figure}[htb]
\centering
\includegraphics[width=0.75\textwidth]{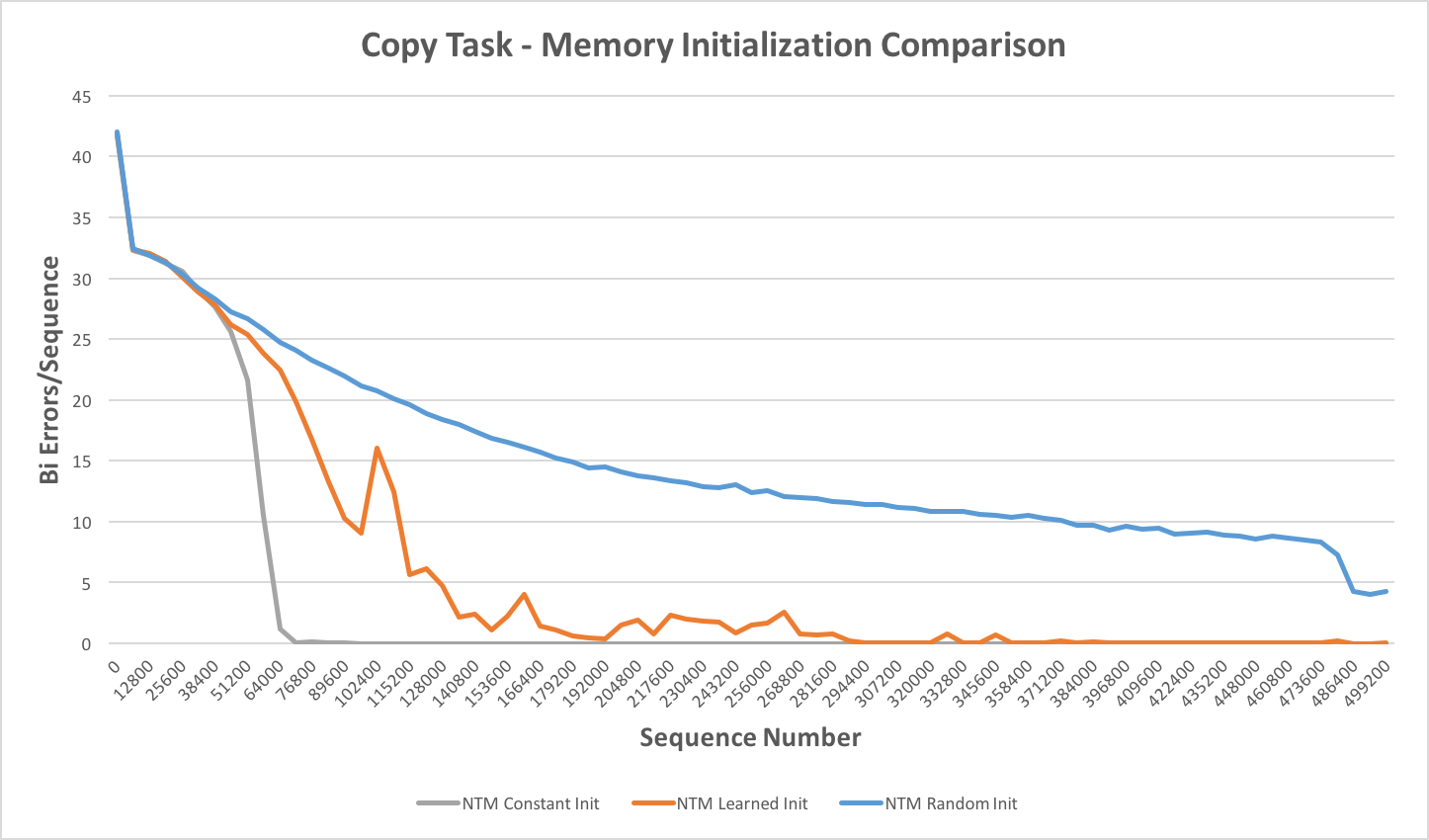}
\caption{Copy task memory initialization comparison - learning curves. Median error on 10 training runs (each) for a NTM initialized according to the \textit{constant}, \textit{learned} and \textit{random} initialization schemes.}
\vspace{-0.1em}
\label{fig:copy_memory_initialization_comparison}
\end{figure}

\textbf{Copy} - Our NTM initialized according the \textit{constant initialization} scheme converges to near zero error approximately 3.5 times faster than the \textit{learned initialization} scheme, while the \textit{random initialization} scheme fails to solve the Copy task in the allotted time (Fig. \ref{fig:copy_memory_initialization_comparison}). The learning curves suggest that initializing the memory contents to small constant values offers a substantial speed-up in convergence over the other two memory initialization schemes for the Copy task.

\begin{figure}[htb]
\centering
\includegraphics[width=0.75\textwidth]{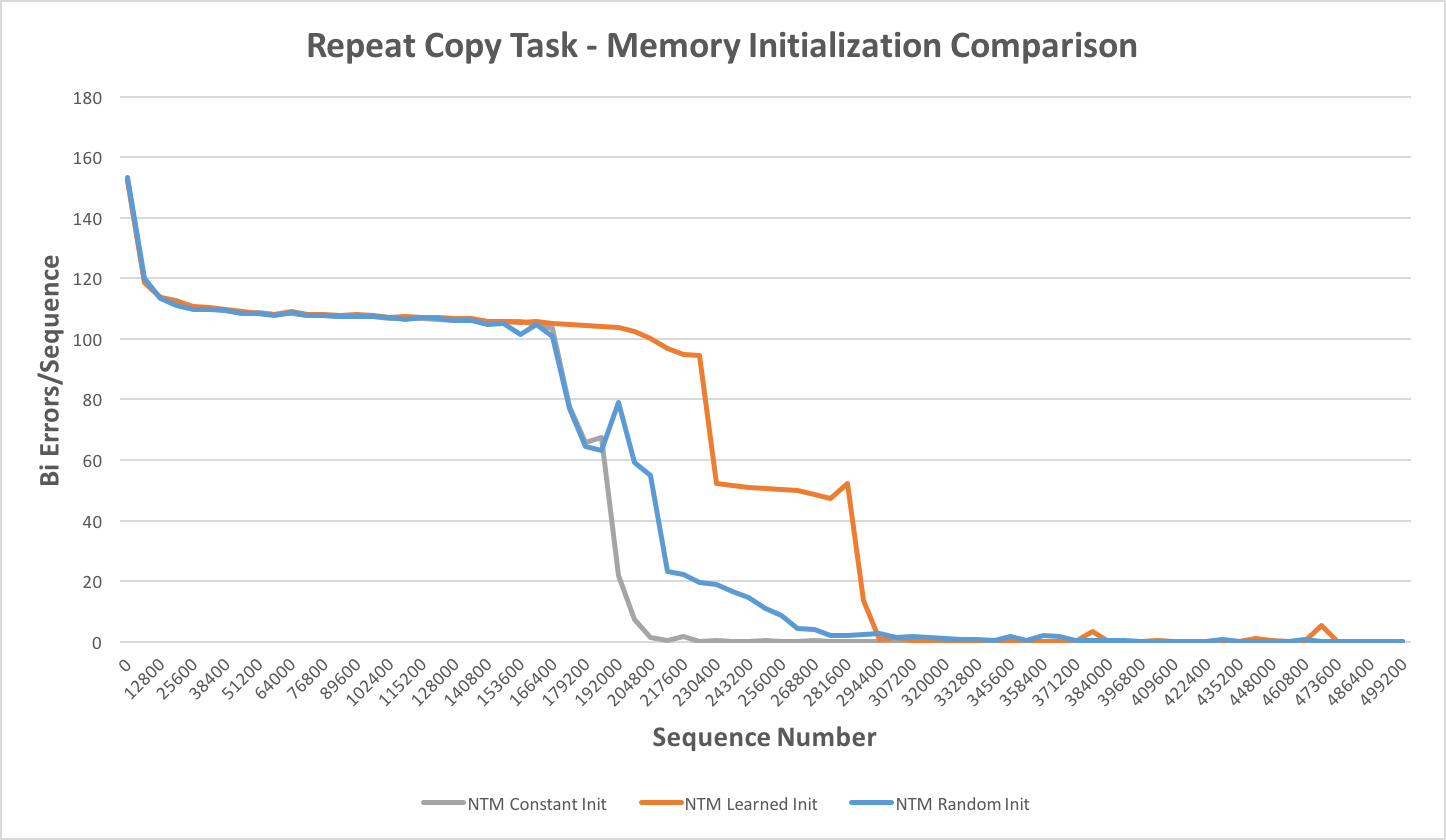}
\caption{Repeat Copy task memory initialization comparison - learning curves. Median error on 10 training runs (each) for a NTM initialized according to the \textit{constant}, \textit{learned} and \textit{random} initialization schemes.}
\vspace{-0.1em}
\label{fig:repeat_copy_memory_initialization_comparison}
\end{figure}

\textbf{Repeat Copy} - A NTM initialized according the \textit{constant initialization} scheme converges to  near zero error approximately 1.43 times faster than the \textit{learned initialization} scheme and 1.35 times faster than the \textit{random initialization} scheme (Fig. \ref{fig:repeat_copy_memory_initialization_comparison}). The relative speed of convergence between \textit{learned} and \textit{random} initialization is reversed as compared with the Copy task, but again the \textit{constant initialization} scheme demonstrates substantially faster learning than either alternative.

\begin{figure}[!htb]
\centering
\includegraphics[width=0.75\textwidth]{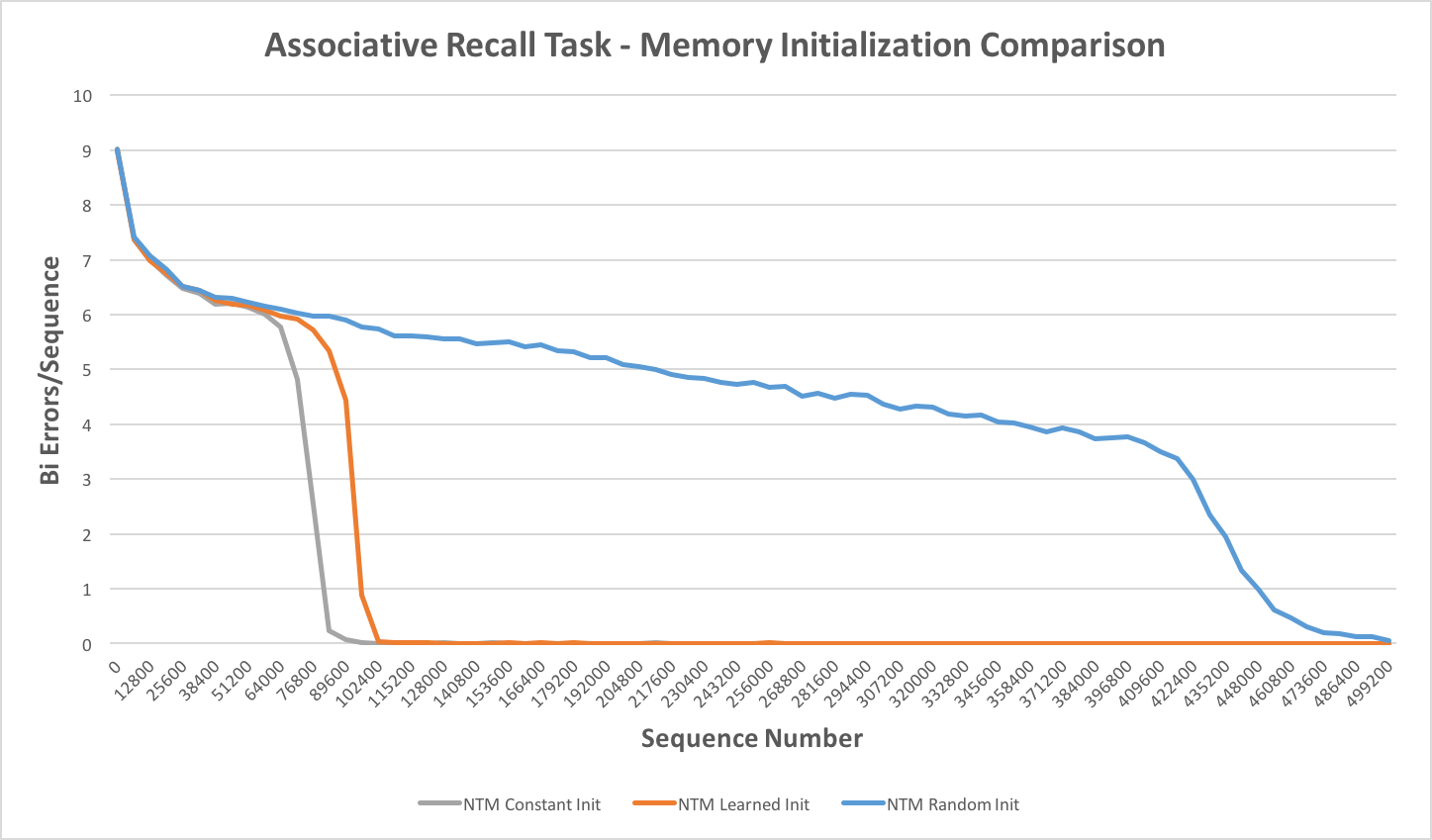}
\caption{Associative Recall task memory initialization comparison - learning curves. Median error on 10 training runs (each) for a NTM initialized according to the \textit{constant}, \textit{learned} and \textit{random} initialization schemes.}
\vspace{-0.1em}
\label{fig:associative_recall_memory_initialization_comparison}
\end{figure}

\textbf{Associative Recall} - A NTM initialized according the \textit{constant initialization} scheme converges to near zero error approximately 1.15 times faster than the \textit{learned initialization} scheme and 5.3 times faster than the \textit{random initialization} scheme (Fig. \ref{fig:associative_recall_memory_initialization_comparison}).

The \textit{constant initialization} scheme demonstrates fastest convergence to near zero error on all three tasks. We conclude that initializing the memory contents of a NTM to small constant values results in faster learning than backpropagating through memory contents initialization or randomly initializing memory contents. Thus, we use the \textit{constant initialization} scheme as the default scheme for our NTM implementation. 

\subsection{Architecture Comparison}

Now that we have established the best memory contents initialization scheme is \textit{constant initialization} we wish to test whether our NTM implementation using this scheme is stable and has similar speed of learning and generalization ability as claimed in the original NTM paper. We compare the performance of our NTM to a LSTM and a DNC on the same three tasks as for our memory contents initialization experiments.

\begin{figure}[htb]
\centering
\includegraphics[width=0.75\textwidth]{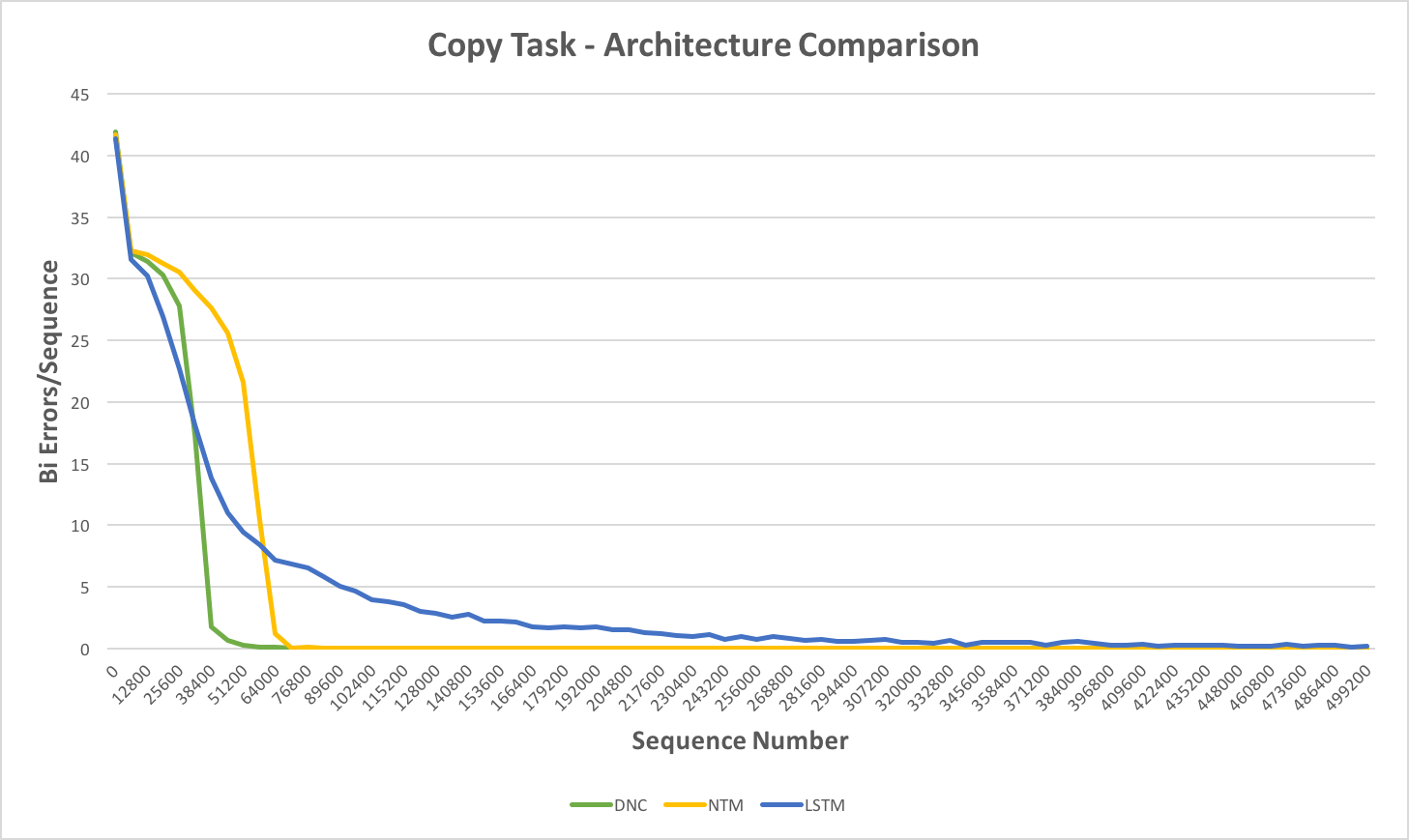}
\caption{Copy task architecture comparison - learning curves. Median error on 10 training runs (each) for a DNC, NTM and LSTM.}
\vspace{-0.1em}
\label{fig:copy_architecture_comparison}
\end{figure}

\textbf{Copy} - Our NTM implementation converges to zero error in a number of steps comparable to the best published results on this task \cite{RN11} (Fig. \ref{fig:copy_architecture_comparison}). Our NTM converges to zero error 1.2 times slower than the DNC and as expected both MANNs learn substantially faster (4-5 times) than a LSTM.

\begin{figure}[htb]
\centering
\includegraphics[width=0.75\textwidth]{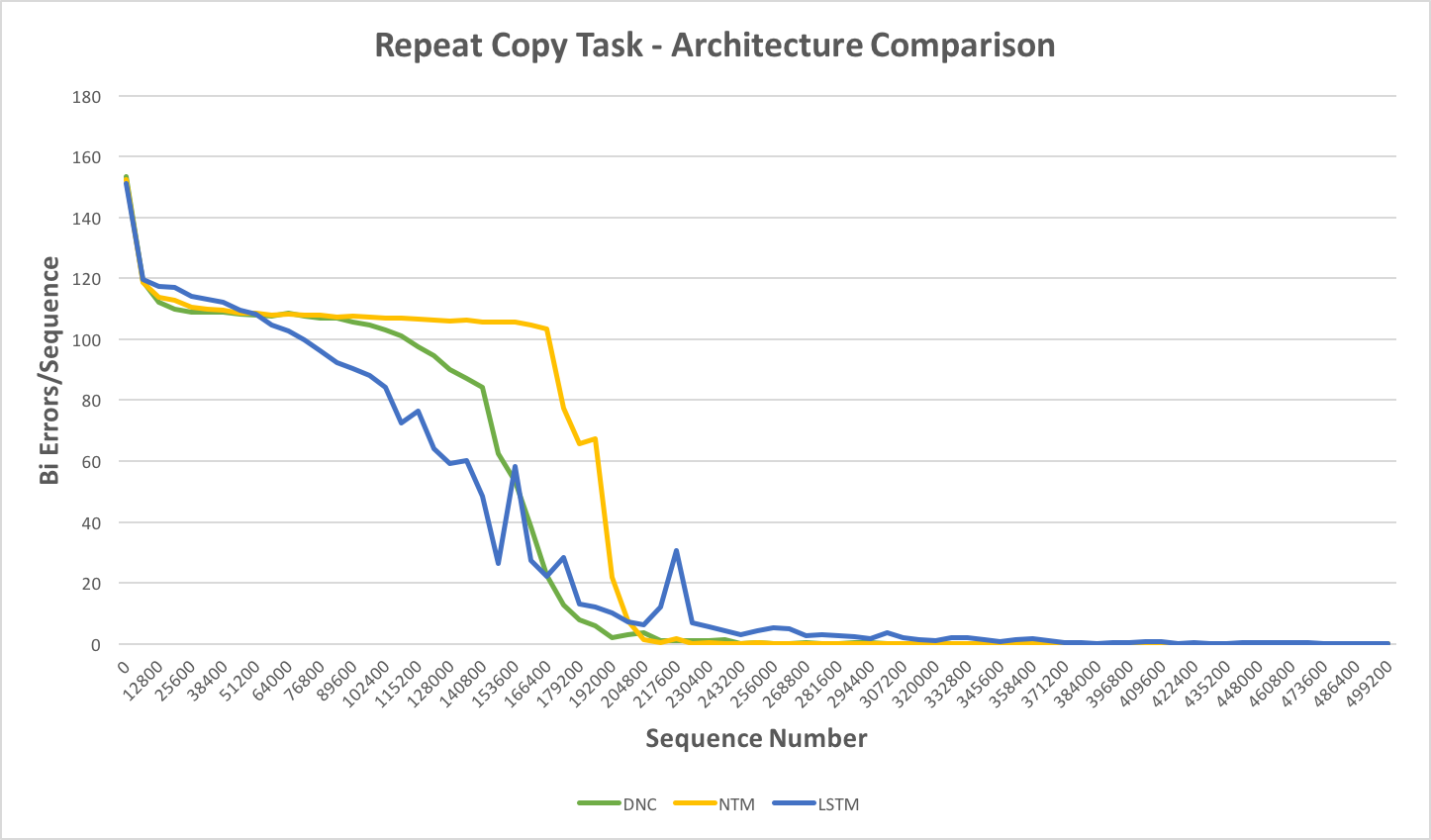}
\caption{Repeat Copy task architecture comparison - learning curves. Median error on 10 training runs (each) for a DNC, NTM and LSTM.}
\vspace{-0.1em}
\label{fig:repeat_copy_architecture_comparison}
\end{figure}

\textbf{Repeat Copy} - As per \cite{RN11}, we also find that the LSTM performs better relative to the MANNs on Repeat Copy compared to the Copy task, converging only 1.44 times slower than a NTM, perhaps due to the shorter input sequences involved (Fig. \ref{fig:repeat_copy_architecture_comparison}). While both the DNC and the NTM demonstrate slow learning during the first third of training both architectures then rapidly fall to near zero error before the LSTM. Despite the NTM learning slower than the DNC during early training, the DNC converges to near zero error just 1.06 times faster than the NTM.

\begin{figure}[!htb]
\centering
\includegraphics[width=0.75\textwidth]{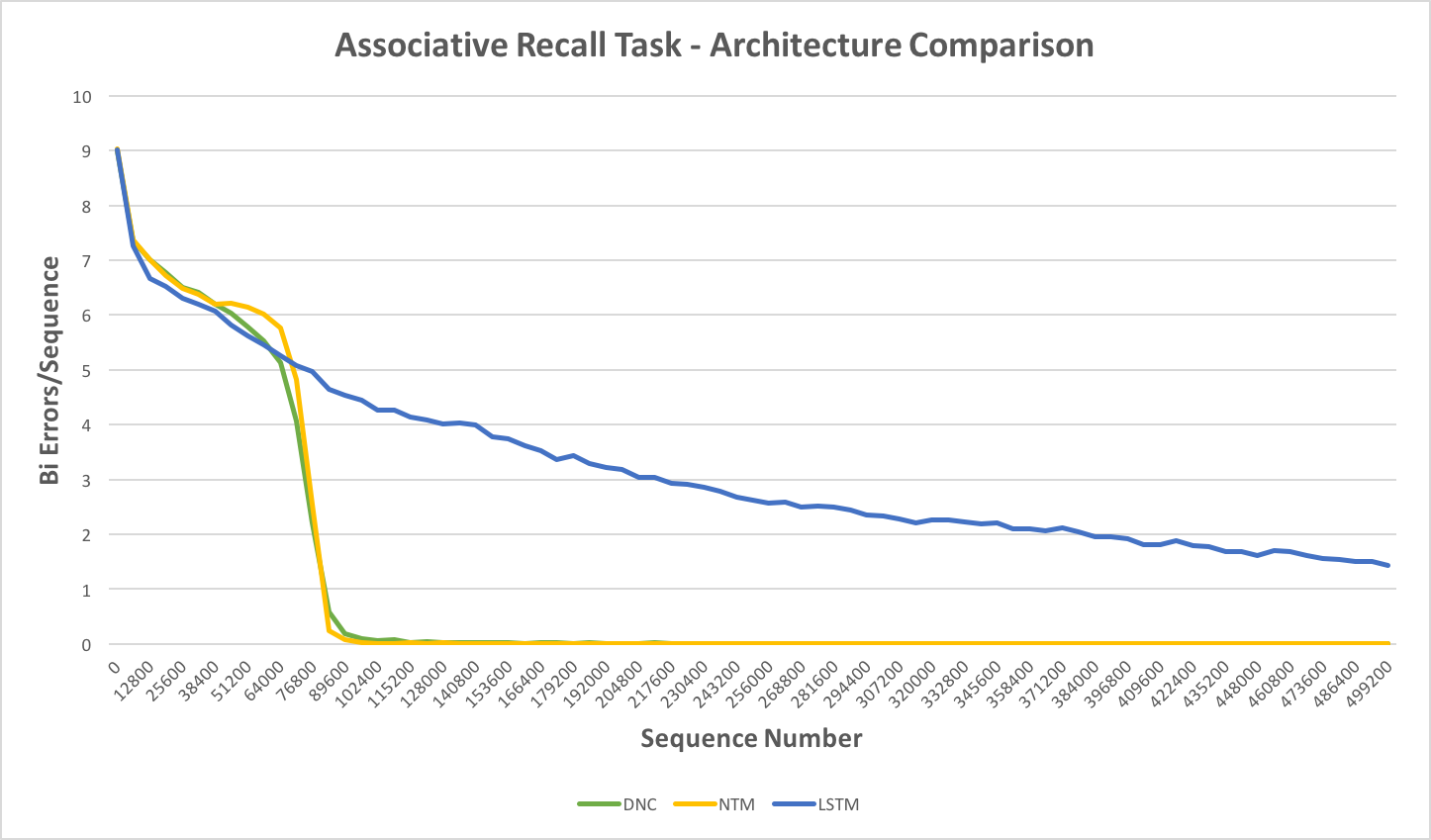}
\caption{Associative Recall task architecture comparison - learning curves. Median error on 10 training runs (each) for a DNC, NTM and LSTM.}
\vspace{-0.1em}
\label{fig:associative_recall_archiecture_comparison}
\end{figure}

\textbf{Associative Recall} - Our NTM implementation converges to zero error in a number of steps almost identical to the best published results on this task \cite{RN11} and at the same rate as the DNC (Fig. \ref{fig:associative_recall_archiecture_comparison}). The LSTM network fails to solve the task in the time provided.

Our NTM implementation learns to solve all three of the five tasks proposed in the original NTM paper \cite{RN11} that we tested. Our implementation's speed to convergence and relative performance to LSTMs is similar to the results reported in the NTM paper. Speed to convergence for our NTM is only slightly slower than a DNC - another MANN. We conclude that our NTM implementation can be used reliably in new applications of MANNs.

\section{Summary} \label{summary}

NTMs are an exciting new neural network architecture that achieve impressive performance on a range of artificial tasks. But the specification of a NTM leaves many free choices to the implementor and no source code is provided that makes these choices and replicates the published NTM results. In practice the choices left to the implementor have a significant impact on the ability of a NTM to learn. We observe great diversity in how these choices are made amongst open source efforts to implement a NTM, many of which fail to replicate these published results.

We have demonstrated that the choice of memory contents initialization scheme is crucial to successfully implementing a NTM. We conclude from the learning curves on three sequential learning tasks that learning is fastest under the \textit{constant initialization} scheme. We note that the \textit{random initialization} scheme which was used in five of the seven identified open source implementations was the slowest to converge on two of the three tasks and the second slowest on the Repeat Copy task.

We have made our NTM implementation with the \textit{constant initialization} scheme open source. Our implementation has learned the Copy, Repeat Copy and Associative Recall tasks at a comparable speed to previously published results and the official implementation of a DNC. Training of our NTM is stable and does not suffer from problems such as gradients becoming NaN reported in other implementations. Our implementation can be reliably used for new applications of NTMs. Additionally, further research on NTMs will be aided by a stable, performant open source NTM implementation.

\subsubsection{Acknowledgements}

This publication emanated from research conducted with the financial support of Science Foundation Ireland (SFI) under Grant Number 13/RC/2106.

\section*{References} \label{references}
\vspace{-1cm}

\bibliographystyle{splncs03}
\bibliography{ntm_implementation}

\begin{thebibliography}{10}
\providecommand{\url}[1]{\texttt{#1}}
\providecommand{\urlprefix}{URL }

\bibitem{RN6}
Bahdanau, D., Cho, K., Bengio, Y.: Neural machine translation by jointly
  learning to align and translate. arXiv preprint arXiv:1409.0473  (2014)

\bibitem{RN36}
Graves, A., Liwicki, M., Fernández, S., Bertolami, R., Bunke, H., Schmidhuber,
  J.: A novel connectionist system for unconstrained handwriting recognition.
  IEEE transactions on pattern analysis and machine intelligence  31(5),
  855--868 (2009)

\bibitem{RN37}
Graves, A., Mohamed, A.r., Hinton, G.: Speech recognition with deep recurrent
  neural networks. In: Acoustics, speech and signal processing (icassp), 2013
  ieee international conference on. pp. 6645--6649. IEEE

\bibitem{RN11}
Graves, A., Wayne, G., Danihelka, I.: Neural turing machines. arXiv preprint
  arXiv:1410.5401  (2014)

\bibitem{RN12}
Graves, A., Wayne, G., Reynolds, M., Harley, T., Danihelka, I.,
  Grabska-Barwińska, A., Colmenarejo, S.G., Grefenstette, E., Ramalho, T.,
  Agapiou, J.: Hybrid computing using a neural network with dynamic external
  memory. Nature  538(7626),  471 (2016)

\bibitem{RN52}
Gulcehre, C., Chandar, S., Cho, K., Bengio, Y.: Dynamic neural turing machine
  with soft and hard addressing schemes. arXiv preprint arXiv:1607.00036
  (2016)

\bibitem{RN18}
Hochreiter, S., Schmidhuber, J.: Long short-term memory. Neural computation
  9(8),  1735--1780 (1997)

\bibitem{RN15}
Kingma, D.P., Ba, J.: Adam: A method for stochastic optimization. arXiv
  preprint arXiv:1412.6980  (2014)

\bibitem{RN10}
Luong, T., Pham, H., Manning, C.D.: Effective approaches to attention-based
  neural machine translation. In: Proceedings of the 2015 Conference on
  Empirical Methods in Natural Language Processing. pp. 1412--1421

\bibitem{pascanu2013difficulty}
Pascanu, R., Mikolov, T., Bengio, Y.: On the difficulty of training recurrent
  neural networks. In: International Conference on Machine Learning. pp.
  1310--1318 (2013)

\bibitem{RN38}
Sukhbaatar, S., Weston, J., Fergus, R.: End-to-end memory networks. In:
  Advances in neural information processing systems. pp. 2440--2448

\bibitem{RN16}
Wu, Y., Schuster, M., Chen, Z., Le, Q.V., Norouzi, M., Macherey, W., Krikun,
  M., Cao, Y., Gao, Q., Macherey, K.: Google's neural machine translation
  system: Bridging the gap between human and machine translation. arXiv
  preprint arXiv:1609.08144  (2016)

\end{thebibliography}




\end{document}